\documentclass[10pt,twocolumn,letterpaper]{article}

\usepackage{cvpr}              % To produce the CAMERA-READY version
\usepackage{float}
\usepackage{tabularx}
\usepackage{nicematrix}
\usepackage[normalem]{ulem}
\usepackage{xcolor}
\usepackage{makecell}

\definecolor{cvprblue}{rgb}{0.21,0.49,0.74}
\usepackage[pagebackref,breaklinks,colorlinks,allcolors=cvprblue]{hyperref}

\def\paperID{} % *** Enter the Paper ID here
\def\confName{CVPR}
\def\confYear{2026}

\title{RS-SSM: Refining Forgotten Specifics in State Space Model for\\Video Semantic Segmentation}

\author{Kai Zhu$^1$, Zhenyu Cui$^2$, Zehua Zang$^3$$^,$$^4$, Jiahuan Zhou$^1$\thanks{Corresponding author}\\
   $^1$Wangxuan Institute of Computer Technology, Peking University, 
   $^2$Tsinghua University, \\
  $^3$Institute of Software Chinese Academy of Sciences, $^4$University of Chinese Academy of Sciences\\
  \small\tt{zhukai2022@ruc.edu.cn}, \tt{cuizhenyu@mail.tsinghua.edu.cn}, \\
  \small\tt{zehua2020@iscas.ac.cn}, \tt{jiahuanzhou@pku.edu.cn}
}

\begin{document}
\maketitle
\begin{abstract}
Recently, state space models have demonstrated efficient video segmentation through linear-complexity state space compression. However, Video Semantic Segmentation (VSS) requires pixel-level spatiotemporal modeling capabilities to maintain temporal consistency in segmentation of semantic objects. While state space models can preserve common semantic information during state space compression, the fixed-size state space inevitably forgets specific information, which limits the models' capability for pixel-level segmentation. To tackle the above issue, we proposed a \textbf{R}efining \textbf{S}pecifics \textbf{S}tate \textbf{S}pace \textbf{M}odel approach (\textbf{RS-SSM}) for video semantic segmentation, which performs complementary refining of forgotten spatiotemporal specifics. Specifically, a Channel-wise Amplitude Perceptron (CwAP) is designed to extract and align the distribution characteristics of specific information in the state space. Besides, a Forgetting Gate Information Refiner (FGIR) is proposed to adaptively invert and refine the forgetting gate matrix in the state space model based on the specific information distribution. Consequently, our RS-SSM leverages the inverted forgetting gate to complementarily refine the specific information forgotten during state space compression, thereby enhancing the model's capability for spatiotemporal pixel-level segmentation. Extensive experiments on four VSS benchmarks demonstrate that our RS-SSM achieves state-of-the-art performance while maintaining high computational efficiency. The code is available at \href{https://github.com/zhoujiahuan1991/CVPR2026-RS-SSM}{https://github.com/zhoujiahuan1991/CVPR2026-RS-SSM}.

\vspace{-15pt}

\end{abstract}
    
\section{Introduction}
\label{sec:intro}

\begin{figure}[t]
  \centering
   \includegraphics[width=1.0\linewidth]{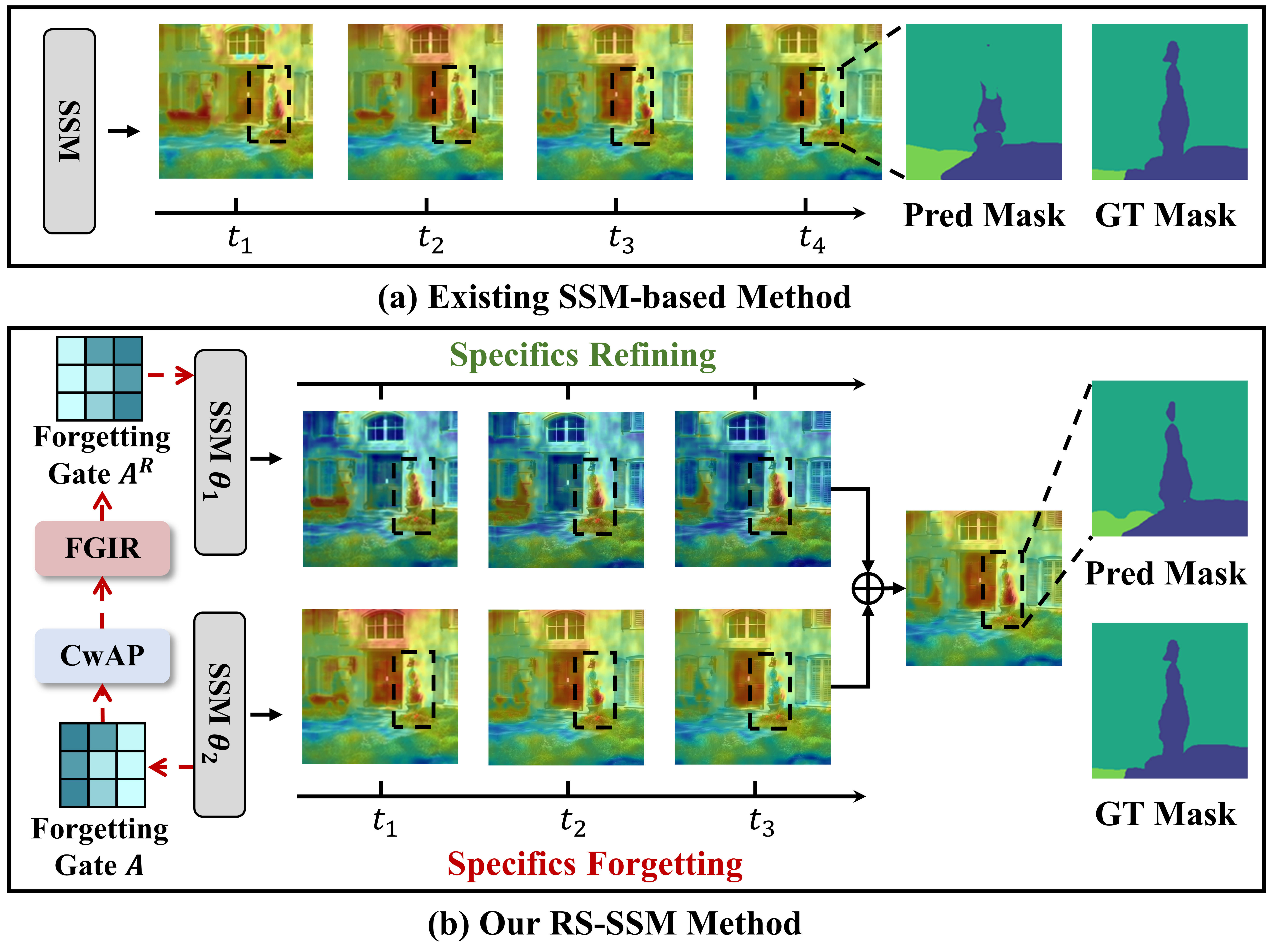}

   \caption{Existing SSM-based VSS methods lose spatiotemporal specifics when perform state space compression, limiting the model's segmentation accuracy. In contrast, our proposed RS-SSM method guides the model to focus on forgotten spatiotemporal specifics, thereby improving the pixel-level semantic segmentation performance.}
   \label{fig:motivation}
\vspace{-20pt}
\end{figure}

Video Semantic Segmentation (VSS) is a fundamental task in the field of computer vision, which aims to assign semantic labels to each pixel in videos and plays a crucial role in understanding the temporal evolution of scenes~\cite{siam2018comparative, siam2021video, ding2019boundary, he2019dynamic, he2019adaptive}. With the advancement of deep learning~\cite{xu2024distribution, xu2025dask, xu2025self, zhou2025distribution, xu2025long, xu2024lstkc, xu2024mitigate, li2024exemplar, cui2024learning, xu2025componential}, VSS has demonstrated extensive practical application potential in areas such as autonomous driving~\cite{guo2024vanishing}, video surveillance~\cite{pierard2023mixture}, and video editing~\cite{xu2022temporally}. In complex long videos, objects may undergo significant displacement, deformation, and occlusion, requiring models to maintain temporal consistency in edge segmentation of semantic objects~\cite{cheng2022masked, qian2024controllable}, which poses critical challenges to their pixel-level spatiotemporal modeling capabilities.

In the field of VSS, early works utilized optical flow to model pixel-level motion information between adjacent frames, propagating and fusing features from previous frames to aggregate temporal information for segmentation of semantic objects~\cite{cheng2017segflow, ding2020every, gao2023exploit, sevilla2016optical, nilsson2018semantic}. However, such methods are not only computationally expensive but also suffer from inaccurate and noisy optical flow estimation in long videos with complex object motion, leading to limited video segmentation accuracy. Recent research has focused on using Transformer-based models to aggregate spatiotemporal features~\cite{hu2020temporally, li2021video, sun2022coarse, sun2022mining, lao2023simultaneously, sun2024learning}. These models leverage the global attention mechanism of the Transformer architecture to aggregate information from multiple frames, improving the segmentation accuracy of semantic objects in videos. However, due to the quadratic complexity of their attention computation, these methods face computational and memory bottlenecks when processing long video sequences.

To efficiently aggregate spatiotemporal information in long video sequences, recent research has introduced methods based on State Space Models (SSM)~\cite{mamba, mamba2, li2024videomamba}. Leveraging selective gating mechanisms, such methods can efficiently compress and propagate spatiotemporal semantic information with fixed-size state space while achieving competitive semantic segmentation accuracy~\cite{hesham2025exploiting}. Nevertheless, as shown in Figure~\ref{fig:motivation}(a), while state space models can preserve common semantic information (e.g., global structure and smooth regions) during state space compression, this fixed-size state space compression process inevitably compromises pixel-level spatiotemporal specifics (e.g., boundaries, textures, local variations), resulting in only coarse segmentation of object locations when performing pixel-level semantic segmentation in videos, but suffering from specifics forgetting. This is detrimental to maintaining temporal consistency of segmentation results for pixel-level semantic information in VSS tasks.

To address the above challenge, we propose a Refining Specifics State Space Model approach, named RS-SSM, which performs complementary refining of the specific information forgotten during the SSM's state space compression process, thereby overcoming the limitations of existing SSM-based VSS methods. Specifically, considering the differences in information contained across the hidden states of different channels in state space models, we design a Channel-wise Amplitude Perceptron (CwAP) module that computes the information distribution characteristics of different channels in the state space model through frequency domain analysis, obtaining Spectrum Features to quantify the differences in specific information amount across channels. Meanwhile, we introduce a channel information loss that minimizes the differences in normalized Spectrum Features across different samples based on cosine similarity, aiming to align the channel distribution characteristics of specific information across samples and encourage RS-SSM to focus on channels rich in specific information. Leveraging the Spectrum Features extracted by the CwAP module, our designed Forgetting Gate Information Refiner (FGIR) module can adaptively invert and refine the forgetting gate matrix of the SSM, thereby guiding our RS-SSM to focus on refining the specific information forgotten during the state space compression process. As shown in Figure~\ref{fig:motivation}(b), benefiting from the guidance of the CwAP and FGIR modules, RS-SSM forms a supplementary specifics refining process, enhancing the capability to capture specific information while maintaining the ability to model common semantic information, thereby improving the pixel-level segmentation performance in VSS tasks.

To sum up, the main contributions of this work are: (1) We identify the limitation of existing SSM-based VSS methods in modeling spatiotemporal specific information and propose a Refining Specifics State Space Model (RS-SSM) to address this issue. (2) We design a Channel-wise Amplitude Perceptron (CwAP) module to extract channel-wise specific information distribution characteristics, and a Forgetting Gate Information Refiner (FGIR) module to adaptively refine the forgetting gate matrix of the SSM, thereby encouraging RS-SSM to focus on spatiotemporal specific information and improving pixel-level segmentation accuracy. (3) Extensive experiments on multiple VSS benchmarks demonstrate that our RS-SSM achieves superior performance against existing methods.

%-------------------------------------------------------------------------

\section{Related Work}
\label{sec:related_work}

\subsection{VSS based on CNNs}

Among all Convolutional Neural Network (CNN)-based methods, Fully Convolutional Networks (FCN) for natural image semantic segmentation pioneered an end-to-end pixel-level classification framework~\cite{shelhamer2017fully}. %and subsequent research improved semantic segmentation accuracy for natural images by introducing dilated convolutions, pyramid structures, and encoder-decoder architectures.
However, as video streams represent a more realistic data modality, researchers have increasingly focused on video semantic segmentation tasks, and how to effectively aggregate and propagate spatiotemporal information in video streams has become a key research focus in this field.

\begin{figure*}[t]
  \centering
   \includegraphics[width=1.0\linewidth]{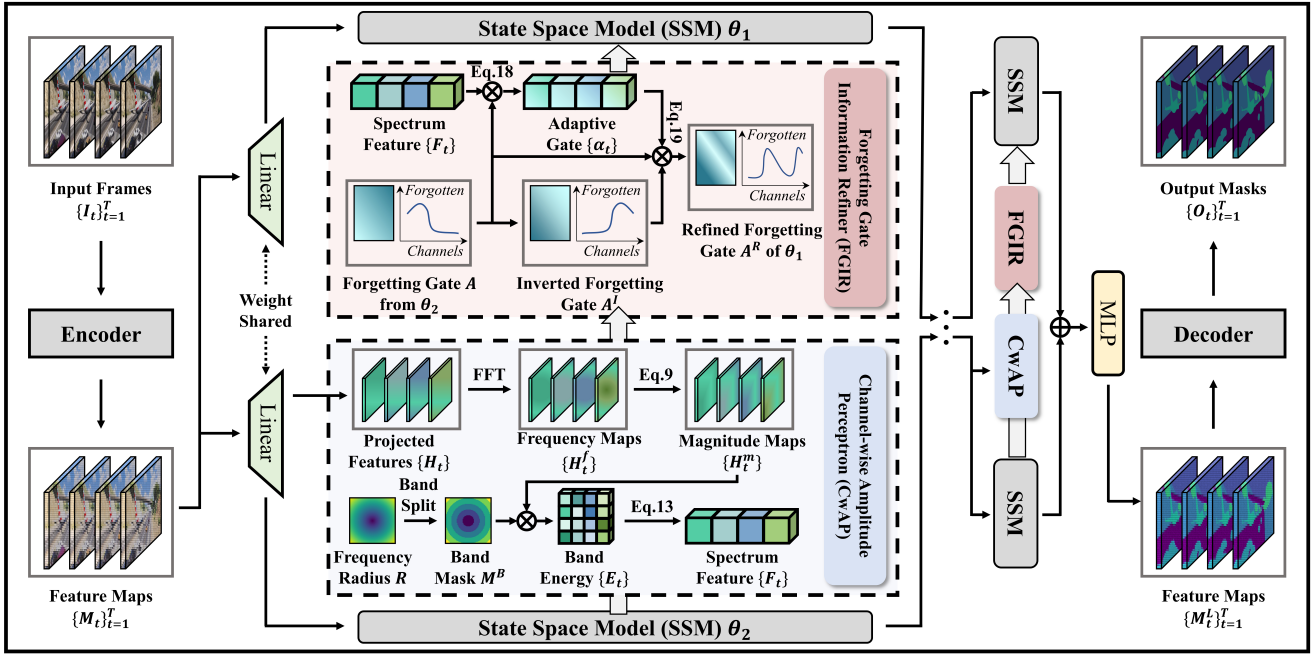}

   \caption{The pipeline of our RS-SSM. Following previous work~\cite{hesham2025exploiting, xie2021segformer}, we use an image encoder to extract feature maps for each frame. After linear projection, we extract spectrum features through the CwAP module to quantify the channel distribution of specific information. Then our FGIR module adaptively inverts and refines the forgetting gate from the SSM \( \theta_2 \), thereby encouraging SSM \( \theta_1 \) to perform complementary refining of forgotten spatiotemporal specifics.}
   \label{fig:method}
\end{figure*}

Early works combined CNNs~\cite{li2018low} with feature propagation based on optical flow estimation and other techniques, leveraging spatiotemporal information to improve segmentation accuracy or reduce computational costs. DFF~\cite{zhu2017deep} and Accel~\cite{jain2019accel} propagate features between adjacent frames through optical flow estimation, while GSVNET~\cite{lee2021gsvnet} and EVS~\cite{paul2020efficient} reduce computational burden by propagating features from key frames to non-key frames.
ClockNet~\cite{shelhamer2016clockwork} and LLVS~\cite{li2018low} introduce adaptive feature reuse, utilizing cross-frame semantic similarity to improve segmentation efficiency and temporal consistency.
However, the uncertainty of dynamic scenes makes the aforementioned methods susceptible to cumulative errors, thereby reducing segmentation accuracy. ETC~\cite{liu2020efficient} encodes multi-frame dependencies as latent embeddings by introducing the recurrent unit ConvLSTM~\cite{shi2015convolutional} to leverage temporal information for improved segmentation accuracy. However, the high computational cost and memory requirements of recurrent networks, as well as LSTM's sensitivity to sequence length~\cite{pfeuffer2019semantic}, limit the scalability of such methods when processing long video sequences.

\subsection{VSS based on Transformers}

Inspired by the success of Transformer models in natural language processing~\cite{vaswani2017attention}, recent researchers have shifted toward using Transformer architectures in VSS to better capture long-range spatiotemporal dependencies in videos~\cite{xu2025dual}. TDNet~\cite{hu2020temporally} uses attention modules to aggregate and propagate features from consecutive frames, while STT~\cite{li2021video} and LMANet~\cite{paul2021local} extract features from reference frames through attention mechanisms to assist in segmenting target frames. MPVSS~\cite{weng2023mask} achieves efficient temporal aggregation for longer video sequences through a memory-augmented Transformer framework. Similarly, CFFM~\cite{sun2022coarse} and MRCFA~\cite{sun2022mining} employ multi-resolution inter-frame attention mechanisms to handle temporal variations through intra-frame contextual static-dynamic decoupling, thereby achieving pixel-level segmentation processes that distinguish between stationary and moving elements.

Although Transformer-based VSS models have achieved significant progress, the quadratic complexity of their attention mechanism results in high computational costs and memory requirements when processing high-resolution frames or long video sequences~\cite{shen2021efficient, katharopoulos2020transformers, wang2020linformer}. This challenge highlights the need for next-generation VSS model architectures that can effectively model long-range spatiotemporal information while improving computational efficiency.

\subsection{State Space Models}

In recent years, State Space Models (SSMs) have emerged as a promising approach for sequence modeling~\cite{S4, dss}. Through HiPPO initialization~\cite{gu2020hippo, gu2022parameterization}, SSMs possess the ability to capture long-range dependencies with linear computational complexity. Building on this foundation, Mamba and Mamba2 introduced input-dependent update and forget gates to address limitations in content-based reasoning~\cite{mamba, mamba2}. Unlike Transformer architectures, this allows SSM parameters to be dynamically modulated by input, significantly improving expressiveness in discrete modalities. Coupled with hardware-friendly parallelization, these advances lead to notable gains in computational efficiency and performance in language modeling~\cite{waleffe2024empirical, ye2025longmamba, zimerman2025explaining}.

Building on this progress, recent work has introduced SSMs into the vision domain~\cite{vim, liu2024vmamba, han2024demystify, li2024videomamba, park2024videomamba, xiao2025spatialmamba, yang2024vivim, ma2024u, yu2025mambaout, zhoustate}. Vision Mamba extends the Mamba framework to 2D image modeling through bidirectional spatial scanning~\cite{vim, liu2024vmamba}, while U-Mamba combines SSMs with convolutional layers, achieving superior performance in medical image segmentation tasks~\cite{ma2024u}. VideoMamba further introduces spatiotemporal scanning, becoming a competitive backbone network in video domain~\cite{li2024videomamba, park2024videomamba, chen2024video, lu2025snakes}. TV3S is the first to apply SSMs to video semantic segmentation tasks, achieving efficient information propagation in long video sequences~\cite{hesham2025exploiting}. However, existing SSM-based VSS methods often overlook the rich spatiotemporal specifics in videos during state space compression, which limits the pixel-level segmentation accuracy of such methods in videos. To this end, our proposed RS-SSM effectively refines the forgotten specifics, achieving superior segmentation performance.

\begin{figure*}[t]
  \centering
   \includegraphics[width=1.0\linewidth]{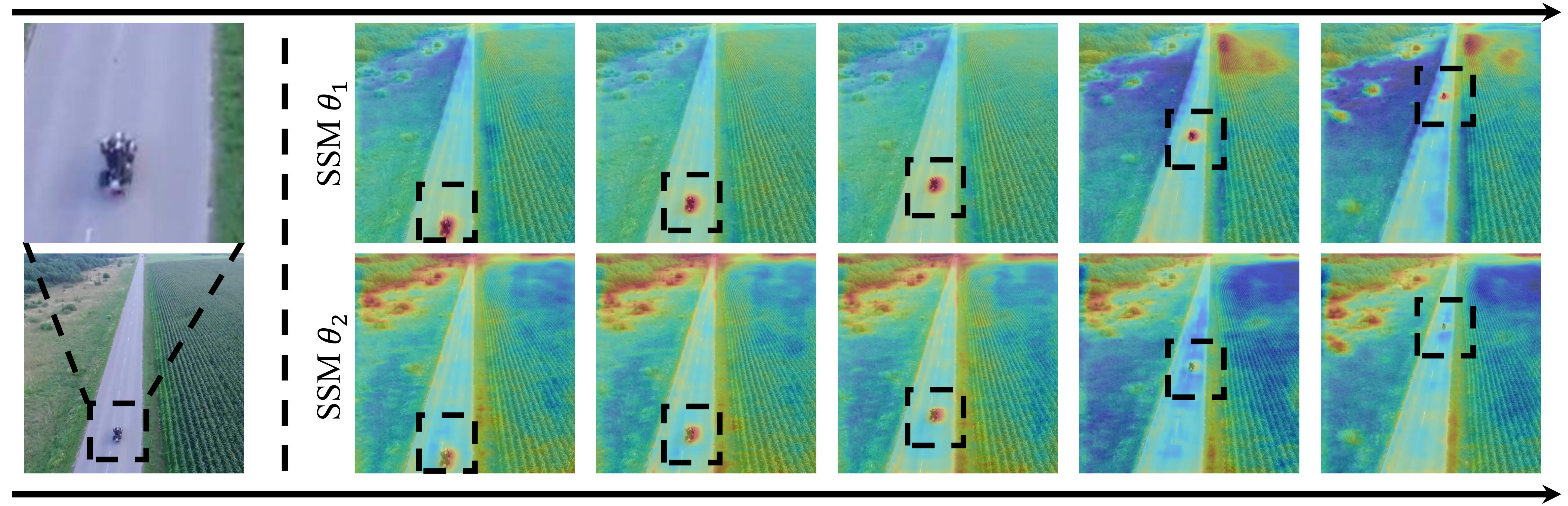}

   \caption{Visualization of the updating gate \( \mathbf{\overline{B}}_d \) as mentioned in Eq~\ref{eq:gate}. Influenced by the forgetting gate, the updating gate reflects the amount of new information introduced at each time step. The bottom row visualization reveals that the vanilla SSM \( \theta_2 \) suffers from information loss of specifics during the state space compression. Conversely, as illustrated in the top row, the SSM \( \theta_1 \) effectively refines these forgotten specifics after inverting and refining the forgetting gate.}
   \label{fig:updating1}
\end{figure*}

\section{Method}
\label{sec:method}

In this section, we illustrate the proposed RS-SSM comprehensively, and the overall pipeline is depicted in Figure~\ref{fig:method}.

\subsection{Preliminary}
\label{sec:preliminary}

The SSM-based models are inspired by continuous systems that map one-dimensional functions \( x(t) \in \mathbb{R} \mapsto y(t) \in \mathbb{R} \) through a hidden state \( h(t) \in \mathbb{R}^{D_s} \), where the state space is \( D_s \)-dimensional.

For multi-channel token processing, the model maintains \( D \) independent SSMs, where each channel \( d \in \{1, \ldots, D\} \) has its own \( D_s \)-dimensional state space. To reduce computational complexity, the state evolution matrix is diagonalized. For a single channel \( d \), the hidden state evolves as follows:
\begin{equation}
\begin{aligned}
h_d^{\prime}(t) = \mathbf{A}_d \odot h_d(t) + \mathbf{B}_d x_d(t), \quad y_d(t) = \mathbf{C}_d^\top h_d(t),
\end{aligned}
\end{equation}
where \( \mathbf{A}_d \in \mathbb{R}^{D_s} \) is the diagonal forgetting vector for channel \( d \), \( \mathbf{B}_d \in \mathbb{R}^{D_s} \) denotes the update vector, \( \mathbf{C}_d \in \mathbb{R}^{D_s} \) serves as the output projection vector, and \( \odot \) represents element-wise multiplication. Across all channels, the forgetting vector forms a forgetting gate matrix \( \mathbf{A} \in \mathbb{R}^{D \times D_s} \).

To facilitate application in deep learning, SSMs are discretized into discrete-time systems. For each channel, the continuous parameters are transformed using a learnable time step \( \Delta_d \in \mathbb{R} \):
\begin{equation}
\label{eq:gate}
\begin{aligned}
&\mathbf{\overline{A}}_d = \exp(\Delta_d \mathbf{A}_d), \\
&\mathbf{\overline{B}}_d = (\Delta_d \mathbf{A}_d)^{-1} \odot (\exp(\Delta_d \mathbf{A}_d) - \mathbf{1}) \odot (\Delta_d \mathbf{B}_d),
\end{aligned}
\end{equation}
where operations are element-wise. The discretized SSM for channel \( d \) is:
\begin{equation}
\begin{aligned}
h_{d,t} = \mathbf{\overline{A}}_d \odot h_{d,t-1} + \mathbf{\overline{B}}_d x_{d,t}, \quad y_{d,t} = \mathbf{C}_d^\top h_{d,t},
\end{aligned}
\end{equation}
where \( h_{d,t-1}, h_{d,t} \in \mathbb{R}^{D_s} \), and \( x_{d,t}, y_{d,t} \in \mathbb{R} \). For a complete token at time \( t \), we have \( \mathbf{x}_t = [x_{1,t}, \ldots, x_{D,t}]^\top \in \mathbb{R}^{D} \) and \( \mathbf{y}_t = [y_{1,t}, \ldots, y_{D,t}]^\top \in \mathbb{R}^{D} \).

The discretized evolution parameters \( \mathbf{\overline{A}}_d = \exp(\Delta_d \mathbf{A}_d) \) act as element-wise decay factors that govern the degree of information compression from previous hidden states. Since \( \mathbf{A}_d \) is constrained to be negative, each element of \( \mathbf{\overline{A}}_d \) lies in \( (0, 1) \), where smaller values impose more aggressive state space compression on historical information in \( h_{d,t-1} \), while values closer to 1 preserve more information with minimal state space compression.

\subsection{Overall Architecture}

Our RS-SSM architecture processes input video frames \( \{I_t\}_{t=1}^T \), where \( T \) is the number of frames. An image encoder \( \mathcal{E}(\cdot) \) extracts feature maps \( \{M_t\}_{t=1}^T \) from each frame:
\begin{equation}
\begin{aligned}
M_t = \mathcal{E}(I_t), \quad t \in \{1, \ldots, T\}.
\end{aligned}
\end{equation}
Each feature map \( M_t \in \mathbb{R}^{C \times H \times W} \) has channel dimension \( C \), height \( H \) and width \( W \). After the image encoder, we stack \( L \) dual-path SSM layers, and each layer contains two SSM modules \( \theta_1^l, \theta_2^l, l \in \{1, \dots, L\} \). For the input to the \( l \)-th dual-path layer, we denote it as \( M^{l-1}_t \). Within each dual-path SSM layer, the input feature map is first mapped to a \( D \)-dimensional projected feature through a weight-shared linear projection \( \mathcal{P( \cdot )} \):
\begin{equation}
\label{eq:shared_linear}
\begin{aligned}
H_t^{l-1} = \mathcal{P}(M^{l-1}_t), \quad t \in \{1, \ldots, T\},
\end{aligned}
\end{equation}
where \( H_t^{l-1} \in \mathbb{R}^{D \times H \times W} \). Subsequently, the projected features \( \{H_t^{l-1}\}_{t=1}^T \) are fed into the SSM modules for spatiotemporal modeling, respectively. These outputs are concatenated along the channel dimension and then fused through an MLP layer to obtain the output feature map of the \( l \)-th layer:
\begin{equation}
\begin{aligned}
M^{l}_t = \text{MLP}(\text{Concat}[\theta_1^{l}(H_t^{l-1}), \theta_2^{l}(H_t^{l-1})]), 
\end{aligned}
\end{equation}
where \(t \in \{1, \ldots, T\}\). Finally, after passing through all \( L \) dual-path SSM layers, the output feature maps \( \{M^{L}_t\}_{t=1}^T \) are fed into a linear segmentation decoder \( \mathcal{D} \) to generate the final segmentation masks for each frame:
\begin{equation}
\begin{aligned}
O_t = \mathcal{D}(M^{L}_t), \quad t \in \{1, \ldots, T\}.
\end{aligned}
\end{equation}

\subsection{Channel-wise Amplitude Perceptron (CwAP)}

As mentioned in Sec~\ref{sec:intro}, the CwAP module computes the information distribution characteristics of different channels in the projected features \( \{H_t\}_{t=1}^T \), where \( H_t \in \mathbb{R}^{D \times H \times W} \). Following previous works~\cite{chen2024frequency, chen2025frequency}, we employ frequency domain transformation to quantify the amount of specific information in different channels. For each projected feature \( H_t \), we first apply 2D Fast Fourier Transform to convert it into the frequency domain:
\begin{equation}
\begin{aligned}
H_t^{f} = \text{FFT2D}(H_t) \in \mathbb{C}^{D \times H \times W}.
\end{aligned}
\end{equation}
To extract the energy distribution information in the frequency domain and thereby quantify the energy proportion across different frequency bands, we compute the magnitude map as follows:
\begin{equation}
\begin{aligned}
H_t^{m} = \sqrt{(\text{Re}(H_t^{f}))^2 + (\text{Im}(H_t^{f}))^2} \in \mathbb{R}^{D \times H \times W},
\end{aligned}
\end{equation}
where \( \text{Re}(\cdot) \) and \( \text{Im}(\cdot) \) denote the real and imaginary parts, respectively. We partition the frequency domain into \( K \) bands based on normalized frequency radius, where low-frequency bands correspond to common semantic information (e.g., global structure, smooth regions) and high-frequency bands correspond to specific information (e.g., boundaries, textures, local variations):
\begin{equation}
\begin{aligned}
R_t(h, w) = \sqrt{\left(\frac{h - H/2}{H}\right)^2 + \left(\frac{w - W/2}{W}\right)^2},
\end{aligned}
\end{equation}
where we center the frequency spectrum. For the \( k \)-th frequency band \( k \in \{0, 1, \ldots, K-1\} \), we define the binary band mask:
\begin{equation}
\begin{aligned}
M^b_{t,k} = 1\left\{\frac{k}{K} \leq R_t(h, w) < \frac{k+1}{K}\right\},
\end{aligned}
\end{equation}
where \( 1\{\cdot\} \) is the indicator function that outputs 1 if the condition is satisfied and 0 otherwise. The energy in band \( k \) for each channel is:
\begin{equation}
\begin{aligned}
E_{t,c,k} = \sum_{h,w} H^m_{t,c,h,w} \cdot M^b_{t,k}(h, w).
\end{aligned}
\end{equation}
The energy distribution is normalized to obtain a probability distribution, and the spectrum features \( F_t \in \mathbb{R}^C \) is computed as the cumulative energy ratio from the top-\( k_h \) high-frequency bands:
\begin{equation}
\label{eq:K}
\begin{aligned}
\tilde{E}_{t,c,k} = \frac{E_{t,c,k}}{\sum_{k'=0}^{K-1} E_{t,c,k'}}, \quad F_{t,c} = \sum_{k=K-k_h}^{K-1} \tilde{E}_{t,c,k},
\end{aligned}
\end{equation}
where \( k_h \) denotes the number of high-frequency bands used for specifics distribution measurement. The obtained spectrum features \( F_t \) reflect the amount of specific information in each channel and will be used to guide the subsequent forgetting gate invert and refinement in the FGIR module.

To promote consistency in subsequent specifics refining, we aim to align the channel distribution of specific information across samples, i.e., guide the specific information in different samples to be distributed in similar channels. Specifically, we normalize the features and compute cosine similarity between all pairs, obtaining the similarity matrix for frame batch \( \mathcal{B} = \{1, 2, \ldots, B \cdot T\} \):
\begin{equation}
\begin{aligned}
\hat{F} = \frac{F}{\|F\|_2}, \quad
\mathbf{S}_{i,j} = \hat{F}_i^\top \hat{F}_j, \quad \forall i, j \in \mathcal{B},
\end{aligned}
\end{equation}
where \( B \) is the batch size of videos, and \( T \) is the number of sampled frames per video.

We introduce a channel information loss \( \mathcal{L}_{\text{ci}} \) to maximize the average pairwise cosine similarity in the batch, defined as:
\begin{equation}
\label{eq:loss_ci}
\begin{aligned}
\mathcal{L}_{\text{ci}} = 1 - \frac{1}{|\mathcal{B}|^2} \sum_{i=1}^{|\mathcal{B}|} \sum_{j=1}^{|\mathcal{B}|} \mathbf{S}_{i,j}.
\end{aligned}
\end{equation}
%By minimizing this loss, we encourage the specific information in different samples to be encoded in similar channels, thereby promoting the consistency and effectiveness of specific information refining.
By minimizing this loss, the model learns to allocate specific information to the same subset of channels. This consistency allows the subsequent FGIR module to selectively enhance these specifics-rich channels, thereby promoting the effectiveness of specifics refining.

\begin{table*}
    \small
    \centering
  \caption{The comparison with existing methods on the VSPW~\cite{miao2021vspw}, NYUv2~\cite{silberman2012indoor} and CamVid~\cite{brostow2008segmentation} dataset. Our RS-SSM achieves state-of-the-art performance while maintaining high efficiency. The best and second-best results are highlighted in \textcolor{red}{\textbf{RED}} and \textcolor{blue}{\textbf{BLUE}}, respectively. We calculate GFLOPs and FPS with an input resolution of 480 \( \times \) 853.}
  
    \setlength{\tabcolsep}{3.5mm}
    \renewcommand{\arraystretch}{0.95}
  \begin{NiceTabular}{l|c|c|c|c|c|c|c}
    \CodeBefore
    \rectanglecolor{gray!20}{7-1}{7-8}
    \rectanglecolor{gray!20}{12-1}{12-8}
    \rectanglecolor{gray!20}{21-1}{21-8}
    \rectanglecolor{gray!20}{25-1}{25-8}
    \rectanglecolor{gray!20}{29-1}{29-8}
    \Body
    \toprule
    \Block{2-1}{Methods} & \Block{2-1}{Backbones} & \Block{1-3}{mIoU$\uparrow$} & & & \Block{2-1}{GFLOPs$\downarrow$} & \Block{2-1}{Params(M)$\downarrow$} & \Block{2-1}{FPS$\uparrow$} \\
    \cline{3-5}
    & & VSPW & NYUv2 & CamVid & & & \\
    \cline{1-8}
    Segformer~\cite{xie2021segformer} & MiT-B1 & 36.5 & 37.6 & 57.9 & 26.6 & 13.8 & 58.7 \\
    CFFM++~\cite{sun2024learning} & MiT-B1 & 39.9 & 39.5 & 60.1 & 114.3 & 16.5 & 27.6 \\
    MRCFA~\cite{sun2022mining} & MiT-B1 & 38.9 & 39.1 & 59.7 & 77.5 & 16.2 & 40.1 \\
    TV3S~\cite{hesham2025exploiting} & MiT-B1 & \textcolor{blue}{\textbf{40.0}} & \textcolor{blue}{\textbf{39.7}} & \textcolor{blue}{\textbf{60.8}} & 36.9 & 17.3 & 24.7 \\
    %\cline{1-8}
    RS-SSM (Ours) & MiT-B1 & \textcolor{red}{\textbf{41.7}} & \textcolor{red}{\textbf{41.4}} & \textcolor{red}{\textbf{62.7}} & 48.1 & 20.2 & 37.3 \\
    \cline{1-8}
    Segformer~\cite{xie2021segformer} & MiT-B2 & 43.9 & 42.1 & 59.8 & 100.8 & 24.8 & 16.2 \\
    CFFM++~\cite{sun2024learning} & MiT-B2 & 45.5 & 43.9 & 61.0 & 172.6 & 28.5 & 21.5 \\
    MRCFA~\cite{sun2022mining} & MiT-B2 & 45.3 & 43.5 & 60.6 & 127.9 & 27.3 & 10.7 \\
    TV3S~\cite{hesham2025exploiting} & MiT-B2 & \textcolor{blue}{\textbf{46.3}} & \textcolor{blue}{\textbf{44.7}} & \textcolor{blue}{\textbf{61.8}} & 53.9 & 28.3 & 21.9 \\
    %\cline{1-8}
    RS-SSM (Ours) & MiT-B2 & \textcolor{red}{\textbf{46.8}} & \textcolor{red}{\textbf{45.9}} & \textcolor{red}{\textbf{63.7}} & 60.7 & 31.2 & 30.9 \\
    \cline{1-8}
    Mask2Former~\cite{cheng2022masked} & R50 & 38.5 & 41.7 & 58.1 & 110.6 & 44.0 & 19.4 \\
    MPVSS~\cite{weng2023mask} & R50 & 37.5 & 40.1 & 59.7 & 38.9 & 84.1 & 33.9 \\
    Mask2Former~\cite{cheng2022masked} & R101 & 39.3 & 41.3 & 58.9 & 141.3 & 63.0 & 16.9 \\
    MPVSS~\cite{weng2023mask} & R101 & 38.8 & 40.4 & 58.8 & 45.1 & 103.1 & 32.3 \\
    %DeepLabv3+ & R101 & 34.7 & 83.2 & 78.2 & 379.0 & 62.7 & 9.2 \\
    %UpNet & R101 & 36.5 & 82.6 & 76.1 & 403.6 & 83.2 & 16.0 \\
    %PSPNet & R101 & 36.5 & 84.2 & 79.6 & 401.8 & 70.5 & 13.8 \\
    %OCRNet & R101 & 36.7 & 84.0 & 79.0 & 361.7 & 58.1 & 14.3 \\
    %TCB & R101 & 37.8 & 87.9 & 84.0 & 1692 & - & - \\
    %ETC & OCRNet & 37.5 & 84.1 & 79.1 & 361.7 & - & - \\
    Segformer~\cite{xie2021segformer} & MiT-B5 & 48.9 & 46.1 & 61.4 & 185.0 & 82.1 & 9.4 \\
    CFFM++~\cite{sun2024learning} & MiT-B5 & \textcolor{blue}{\textbf{50.1}} & 46.7 & \textcolor{blue}{\textbf{62.3}} & 444.3 & 87.5 & 10.4 \\
    MRCFA~\cite{sun2022mining} & MiT-B5 & 49.9 & 46.7 & 61.8 & 373.0 & 84.5 & 5.0 \\
    TV3S~\cite{hesham2025exploiting} & MiT-B5 & 49.8 & \textcolor{blue}{\textbf{46.9}} & 62.1 & 137.0 & 85.6 & 14.0 \\
    %\cline{1-8}
    RS-SSM (Ours) & MiT-B5 & \textcolor{red}{\textbf{51.6}} & \textcolor{red}{\textbf{49.1}} & \textcolor{red}{\textbf{64.4}} & 128.8 & 88.4 & 20.2 \\
    \cline{1-8}
    Mask2Former~\cite{cheng2022masked} & Swin-T & 41.2 & 37.4 & 58.7 & 114.4 & 47.4 & 17.1 \\
    MPVSS~\cite{weng2023mask} & Swin-T & 39.9 & 36.8 & 58.3 & 39.7 & 114.0 & 32.8 \\
    TV3S~\cite{hesham2025exploiting} & Swin-T & \textcolor{blue}{\textbf{44.9}} & \textcolor{blue}{\textbf{40.5}} & \textcolor{blue}{\textbf{59.4}} & 57.3 & 31.7 & 22.9 \\
    %\cline{1-8}
    RS-SSM (Ours) & Swin-T & \textcolor{red}{\textbf{46.1}} & \textcolor{red}{\textbf{42.2}} & \textcolor{red}{\textbf{60.9}} & 73.5 & 34.6 & 26.7 \\
    \cline{1-8}
    Mask2Former~\cite{cheng2022masked} & Swin-S & 42.1 & 43.1 & 59.6 & 152.2 & 68.9 & 14.5 \\
    MPVSS~\cite{weng2023mask} & Swin-S & 40.4 & 41.7 & 59.3 & 47.3 & 108.0 & 30.6 \\
    TV3S~\cite{hesham2025exploiting} & Swin-S & \textcolor{blue}{\textbf{50.6}} & \textcolor{blue}{\textbf{45.2}} & \textcolor{blue}{\textbf{60.2}} & 94.1 & 53.1 & 19.5 \\
    %\cline{1-8}
    RS-SSM (Ours) & Swin-S & \textcolor{red}{\textbf{52.2}} & \textcolor{red}{\textbf{47.0}} & \textcolor{red}{\textbf{62.0}} & 110.3 & 55.9 & 24.2 \\
    \bottomrule
  \end{NiceTabular}
  \label{tab:comparison1}
\end{table*}

\subsection{Forgetting Gate Information Refiner (FGIR)}

As mentioned in Sec~\ref{sec:preliminary}, the forgetting gate matrix \( \mathbf{A} \in \mathbb{R}^{D \times D_s} \) in SSMs controls the degree of information compression from previous hidden states. In our RS-SSM, we propose the FGIR module to adaptively invert and refine the forgetting gate from the SSM \( \theta_2^l \) based on the spectrum features \( \{F_i\}_{i=1}^{B \cdot T} \) extracted by the CwAP module, and then feed the refined forgetting gate into the SSM \( \theta_1^l \) as its parameter matrix. This encourages RS-SSM to perform complementary refining of specific information.

Given the forgetting gate matrix \( \mathbf{A} \) in the SSM \( \theta_2^l \), we first compute the inverted version of \( \mathbf{A} \) by inverting its value range along the channel dimension:
\begin{equation}
\label{eq:invert}
\begin{aligned}
\mathbf{A}^I = \mathbf{A}_{\max} + \mathbf{A}_{\min} - \mathbf{A} \in \mathbb{R}^{ D \times D_s },
\end{aligned}
\end{equation}
where \( \mathbf{A}_{\max} \) and \( \mathbf{A}_{\min} \) are the maximum and minimum values of \( \mathbf{A} \) along the channel dimension, respectively. This operation inverts the compression degree of the forgetting gate across different channels, making originally highly compressed channels less compressed after inversion, and vice versa.

To quantify the compression degree for different channels in the SSM \( \theta_2^l \), we compute the importance of each channel from \( \mathbf{A} \):
\begin{equation}
\begin{aligned}
\beta_i = \frac{\|\exp((\mathbf{A})_i)\|_2}{\max_j \|\exp((\mathbf{A})_j)\|_2 + \epsilon}, \quad i = 1, \ldots, D,
\end{aligned}
\end{equation}
where \( \beta \in \mathbb{R}^{D} \) represents the normalized channel importance, and \( \epsilon \) is a small constant for the numerical stability. Channels with higher \( \beta_i \) values are less compressed in the SSM \( \theta_2^l \) and thus deemed more important.

The inverting weight \( \alpha \) is computed based on both the spectrum features and channel importance:
\begin{equation}
\begin{aligned}
\tilde{F} &= \text{Softmax}(\frac{1}{B \cdot T} \sum_{i=1}^{B \cdot T} F_i) \in \mathbb{R}^{D}, \\
\alpha &= \tilde{F} \odot (1 - \beta)\in \mathbb{R}^{D}, \\
\end{aligned}
\end{equation}
where \( \odot \) denotes element-wise multiplication. Channels with high specific content (\( \tilde{F} \)) and low importance (\( 1 - \beta \)) in the SSM \( \theta_2^l \) receive higher inverting weights.

Finally, we obtained the refined forgetting gate \( \mathbf{A}^R \) through adaptive interpolation:
\begin{equation}
\begin{aligned}
\mathbf{A}^R = (1 - \alpha) \odot \mathbf{A} + \alpha \odot \mathbf{A}^I,
\end{aligned}
\end{equation}
where \( \mathbf{A}^R \) serves as the forgetting gate in the SSM \( \theta_1^l \). By inverting and refining the forgetting gate from the SSM \( \theta_2^l \), we encourage \( \theta_1^l \) to focus on complementarily refining specific information that is forgotten by \( \theta_2^l \), as shown in Fig~\ref{fig:updating1}, thereby enhancing the overall representation capability of our RS-SSM.

\subsection{Overall Optimization}

Following previous SSM-based VSS works~\cite{hesham2025exploiting}, we employ the standard cross-entropy loss \( \mathcal{L}_{\text{ce}} \) for supervision. Additionally, we incorporate the channel information loss \( \mathcal{L}_{\text{ci}} \) from the CwAP module to promote a consistent distribution of specific information  across samples. The overall loss function is defined as:
\begin{equation}
\label{eq:total_loss}
\begin{aligned}
\mathcal{L} = \mathcal{L}_{\text{ce}}(O_T, Y_T) + \lambda \sum^T_{t=1}\mathcal{L}_{\text{ce}}(O_t, Y_t) + \lambda_i \mathcal{L}_{\text{ci}},
\end{aligned}
\end{equation}
where \( Y_t \) is the ground-truth segmentation mask for frame \( t \). \( \lambda \) and \( \lambda_i \) are hyperparameters that balance the contributions of different loss terms.

\begin{figure*}[t]
  \centering
   \includegraphics[width=1.0\linewidth]{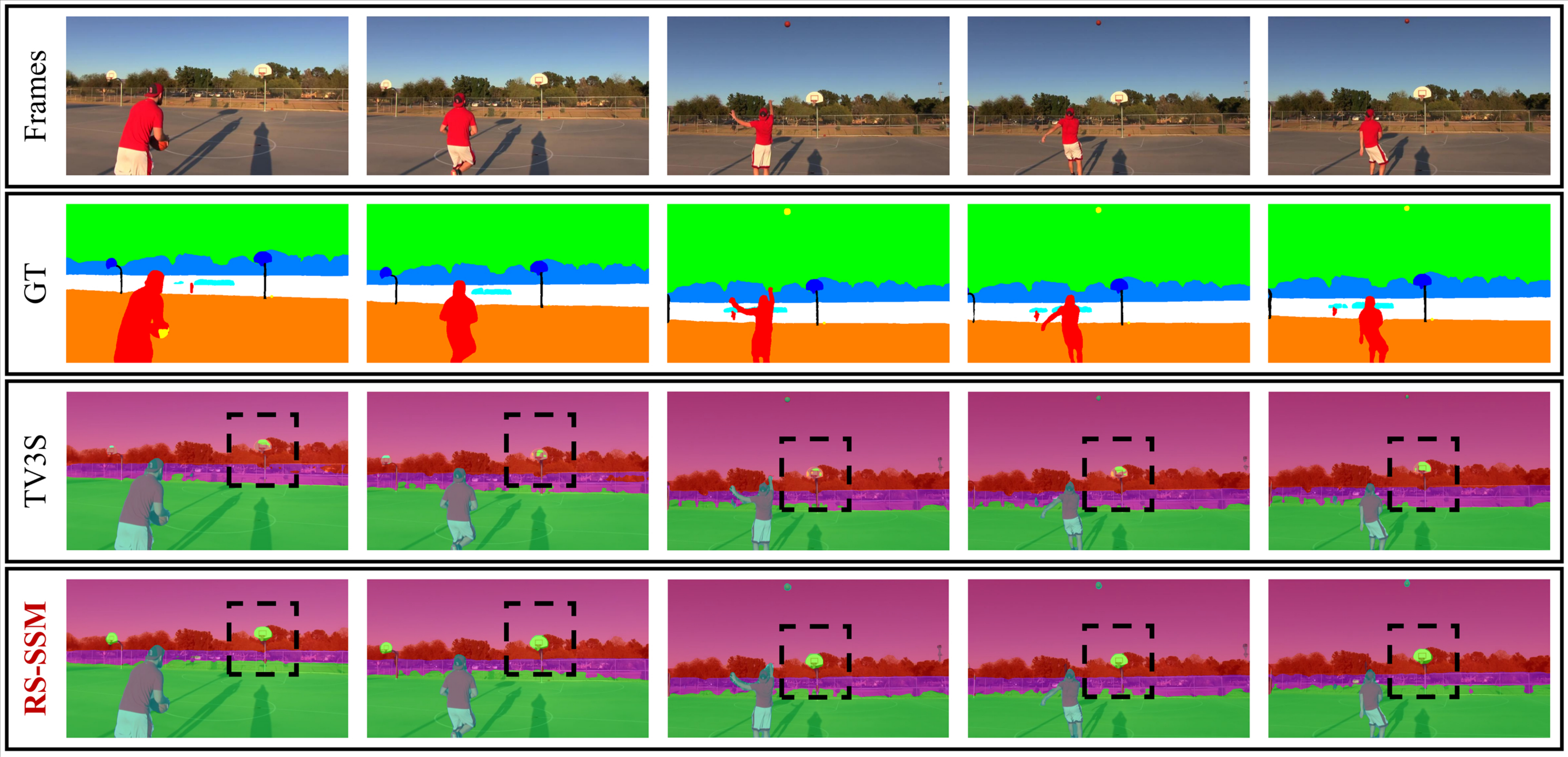}

   \caption{Visualization of segmentation results on VSPW dataset~\cite{miao2021vspw}. Compared to the existing SSM-based method TV3S~\cite{hesham2025exploiting}, our RS-SSM produces more accurate and detailed segmentation results by effectively refining specific information in videos.}
   \label{fig:segres1}
\vspace{-7pt}
\end{figure*}

\section{Experiment}
\label{sec:experiment}

\begin{table}
  \caption{The comparison with existing methods on the Cityscapes~\cite{cordts2016cityscapes} dataset, using an input size of 512 \( \times \) 1024.}
  \label{tab:comparison2}
  \centering
  \footnotesize  % 缩小字体
  \setlength{\tabcolsep}{3pt}  % 减小列间距
  \begin{NiceTabular}{l|c|c|c|c|c}
    \CodeBefore
    \rectanglecolor{gray!20}{9-1}{9-6}
    \Body
    \toprule
    Methods & Backbones & mIoU$\uparrow$ & GFLOPs$\downarrow$ & Params(M)$\downarrow$ & FPS$\uparrow$ \\
    \midrule
    %ETC & R18 & 71.1 & 434.1 & - & - \\
    SegFormer~\cite{xie2021segformer} & MiT-B0 & 71.9 & 54.6 & 3.7 & 58.5 \\
    CFFM~\cite{sun2022coarse} & MiT-B0 & 74.0 & 80.7 & 4.6 & 15.8 \\
    MRCFA~\cite{sun2022mining} & MiT-B0 & 72.8 & 77.5 & 4.2 & 16.6 \\
    \midrule
    SegFormer~\cite{xie2021segformer} & MiT-B1 & 74.1 & 106.4 & 13.7 & 46.8 \\
    CFFM++~\cite{sun2024learning} & MiT-B1 & \textcolor{blue}{\textbf{75.7}} & 169.8 & 15.9 & 20.4 \\
    MRCFA~\cite{sun2022mining}  & MiT-B1 & 75.1 & 145.0 & 14.9 & 13.0 \\
    TV3S~\cite{hesham2025exploiting} & MiT-B1 & 75.6 & 83.6 & 17.3 & 25.1 \\
    RS-SSM (Ours) & MiT-B1 & \textcolor{red}{\textbf{78.3}} & 59.5 & 20.1 & 27.2 \\
    \bottomrule
  \end{NiceTabular}
\end{table}

\begin{table}
  \caption{Ablation study on the influence of different components in our RS-SSM. V-SSM represents only using one vanilla State Space Model path, Bi-V-SSM represents using two-way vanilla State Space Model paths, and No-CwAP using our RS-SSM with a simply inverted forgetting gate as mentioned in Eq~\ref{eq:invert}.}
  \label{tab:ablation1}
  \centering
  \small  % 缩小字体
  \setlength{\tabcolsep}{3pt}  % 减小列间距
  \begin{NiceTabular}{l|c|c|c|c|c}
    \CodeBefore
    \rectanglecolor{gray!20}{6-1}{6-6}
    \Body
    \toprule
    \Block{2-1}{Methods} & \Block{1-5}{mIoU$\uparrow$} \\
    \cline{2-6}
    & MiT-B1 & MiT-B2 & MiT-B5 & Swin-T & Swin-S \\
    \cline{1-6}
    V-SSM & 38.8 & 45.1 & 48.5 & 43.3 & 49.8 \\
    Bi-V-SSM & 39.6 & 45.4 & 49.0 & 43.7 & 50.8 \\
    No-CwAP & 40.7 & 46.0 & 50.2 & 44.9 & 51.6 \\
    RS-SSM & \textcolor{red}{\textbf{41.7}} & \textcolor{red}{\textbf{46.8}} & \textcolor{red}{\textbf{51.6}} & \textcolor{red}{\textbf{45.7}} & \textcolor{red}{\textbf{52.2}} \\
    \bottomrule
  \end{NiceTabular}
\end{table}

\subsection{Datasets and Evaluation Metrics}

The experiments are conducted on four VSS datasets: VSPW~\cite{miao2021vspw}, NYUv2~\cite{silberman2012indoor}, CamVid~\cite{brostow2008segmentation} and Cityscapes~\cite{cordts2016cityscapes}. The largest VSPW~\cite{miao2021vspw} dataset contains 3536 video sequences and covers various scenarios, with 2806 sequences for training, 343 for validation, and 387 for testing, including 124 classes. The NYUv2~\cite{silberman2012indoor} dataset consists of 1449 video sequences captured in indoor environments, with 795 sequences for training and 654 for testing, annotated with 40 classes. The CamVid~\cite{brostow2008segmentation} dataset contains 701 video sequences recorded in urban driving scenarios, with 369 sequences for training, 100 for validation, and 232 for testing, annotated with 32 classes. The Cityscapes~\cite{cordts2016cityscapes} dataset comprises 5000 video sequences captured in urban street scenes, with 2975 sequences for training, 500 for validation and 1525 for testing, annotated with 19 classes.

We use the mean Intersection over Union (mIoU) as the evaluation metric for all datasets. To verify the efficiency of our proposed method, we also report the number of parameters (Params), floating point operations (GFLOPs) and inference speed (FPS) during evaluation.

\subsection{Implementation details}

Our proposed approach is implemented based on MMSegmentation framework and all the experiments were conducted with 4 RTX4090 NVIDIA GPUs. Following the SegFormer~\cite{xie2021segformer}, we utilize MiT and Swin-Transformer~\cite{liu2021swin} as our backbones, which are pre-trained with ImageNet. For fair comparison, we maintain consistency with previous SSM-based VSS methods~\cite{hesham2025exploiting} by using only the current frame \( T \) and frames \( T-9 \), \( T-6 \), and \( T-3 \) as input during the training phase on the VSPW~\cite{miao2021vspw} dataset. And the input embedding dimension is set to 256. Frame inputs for the VSPW~\cite{miao2021vspw} and the CamVid~\cite{brostow2008segmentation} dataset are cropped to 480 \( \times \) 480, while for NYUv2~\cite{silberman2012indoor} and Cityscapes~\cite{cordts2016cityscapes} datasets, frames are cropped to 480 \( \times \) 640 and 512 \( \times \) 1024, respectively. During training, we use the AdamW optimizer with an initial learning rate of \( 6 \times 10^{-5} \) and a weight decay of 0.01. The learning rate is adjusted using the poly learning rate policy with a power of 1.0. Following the prior works, standard data augmentation techniques such as cropping and flipping are applied during training.

\subsection{Comparison with State-of-the-arts}

We compare our proposed RS-SSM with several state-of-the-art VSS methods on VSPW~\cite{miao2021vspw}, NYUv2~\cite{silberman2012indoor}, and CamVid~\cite{brostow2008segmentation}, as shown in Tab~\ref{tab:comparison1}. Among existing methods, TV3S~\cite{hesham2025exploiting} is currently the best SSM-based video semantic segmentation method. We divide the tables into multiple groups based on different backbone types or scales. When using Swin-S and MiT-B5 as backbones, our method surpasses the current state-of-the-art by \textbf{1.6} and \textbf{1.5} mIoU on the large-scale VSPW dataset, respectively, while maintaining efficient inference speed. This is because our RS-SSM design can effectively capture spatiotemporal specifics in videos with linear complexity, enhancing the model's pixel-level segmentation capability for semantic objects, as shown in Fig~\ref{fig:segres1}. When using smaller-scale backbones such as MiT-B1 and Swin-T, our method also outperforms the current state-of-the-art TV3S~\cite{hesham2025exploiting} by \textbf{1.7} and \textbf{1.2} mIoU on the \textbf{VSPW} dataset, respectively. Even when constrained by small-scale backbones, our RS-SSM still demonstrates effective performance improvement.

On the smaller-scale datasets \textbf{NYUv2} and \textbf{CamVid}, our method also achieves significant improvements over existing approaches. When using MiT-B5 as the backbone, our method outperforms the current SSM-based VSS method TV3S~\cite{hesham2025exploiting} by \textbf{2.2} mIoU on the NYUv2 dataset and surpasses the state-of-the-art method CFFM++~\cite{sun2024learning} by \textbf{2.1} mIoU on the CamVid dataset. These results further demonstrate the superior capability of RS-SSM for pixel-level semantic segmentation tasks.

We also evaluate our RS-SSM on the \textbf{Cityscapes}~\cite{cordts2016cityscapes} dataset, as shown in Tab~\ref{tab:comparison2}. When using MiT-B1 as the backbone, our method surpasses the current state-of-the-art by \textbf{2.6} mIoU, demonstrating the effectiveness of RS-SSM in refining spatiotemporal specifics for pixel-level segmentation tasks. This further validates the generalization capability of our RS-SSM across different datasets and scenarios.

Notably, our method not only achieves superior segmentation performance but also exhibits high computational efficiency. As illustrated in Tab~\ref{tab:comparison1}, when employing MiT-B5 as the backbone, our method reduces GFLOPs by \textbf{8.2} and improves FPS by \textbf{6.2} compared to the existing SSM-based VSS method TV3S~\cite{hesham2025exploiting} with an input resolution of 480 \( \times \) 853. As shown in Tab~\ref{tab:comparison2}, when processing input frames of size 512 \( \times \) 1024 pixels with MiT-B1 as the backbone, our method achieves a reduction of \textbf{24.1} in GFLOPs and an improvement of \textbf{2.1} in FPS compared to TV3S~\cite{hesham2025exploiting}. This efficiency gain stems from the design where the linear projection mentioned in Eq~\ref{eq:shared_linear} is weight-shared and the modules CwAP and FGIR are lightweight, thus preserving the computational efficiency inherent to SSM-based VSS methods.

\subsection{Ablation}

To validate the effectiveness of different components in our RS-SSM, we conduct ablation studies on the VSPW dataset, as shown in Tab~\ref{tab:ablation1}. When using only one or two-way vanilla State Space Model (V-SSM and Bi-V-SSM), the performance decreases by up to \textbf{3.1} and \textbf{2.6} mIoU respectively, demonstrating the importance of modeling specific information. When using RS-SSM with a simply inverted forgetting gate (No-CwAP) in Eq~\ref{eq:invert}, the performance decreases by up to \textbf{1.4} mIoU compared to our full RS-SSM. This indicates that the adaptive refinement of the forgetting gate through CwAP is crucial for effectively capturing complementary specific information, thereby enhancing the overall representation capability of our RS-SSM.

\section{Conclusion}
\label{sec:conclusion}
In this paper, we propose a novel Refining Specifics State Space Model (RS-SSM) for video semantic segmentation. Specifically, we design a Channel-wise Amplitude Perceptron (CwAP) module to extract specific information distribution characteristics and align them in the channel-wise from the state space, as well as Forgetting Gate Information Refiner (FGIR) module to adaptively refine the forgetting gate for better refining complementary spatiotemporal specifics. This approach enhances the model's ability to capture pixel-level semantic specifics in videos while maintaining computational efficiency. The effectiveness of our proposed RS-SSM has been validated on four datasets for video semantic segmentation.

\section{Acknowledgements}
\label{sec:acknowledgements}

This work is a research achievement of the National Engineering Research Center of New Electronic Publishing Technologies, supported by the National Natural Science Foundation of China (62376011) and the National Key R\&D Program of China (2024YFA1410000).
{
    \small
    %\clearpage
    \bibliographystyle{ieeenat_fullname}
    \bibliography{main}
}


\end{document}